\documentclass[11pt]{article}

\usepackage[final]{acl}

\usepackage{times}
\usepackage{latexsym}
\usepackage{amssymb}
\usepackage{booktabs}
\usepackage{multirow}
\usepackage{amsmath} 
\usepackage{makecell}   
\usepackage[table]{xcolor}
\usepackage[T1]{fontenc}
\usepackage{graphicx}
\usepackage{xspace}
\usepackage[bb=boondox,bbscaled=.95,cal=boondoxo]{mathalfa}
\newcommand{\X}{$\mathbb{X}$\xspace}

\newcommand{\res}[3]{#1 $\mid$ \textcolor{blue}{#2} $\mid$ \textcolor{purple}{#3}}

\usepackage[utf8]{inputenc}

\usepackage{microtype}

\usepackage{inconsolata}

\usepackage{graphicx}
\usepackage{subfig}
\title{CoVer: Conflict-Aware Claim Verification}

\author{
  \textbf{Shuning Zhang}\textsuperscript{1,*},
  \textbf{Dai Shi}\textsuperscript{2,*},
  \textbf{Bohao Chu}\textsuperscript{3},
  \textbf{Hui Wang}\textsuperscript{3},
  \textbf{Yuwei Chuai}\textsuperscript{4,$\dagger$},\\
  \textbf{Yifan Wang}\textsuperscript{5},
  \textbf{Jingruo Chen}\textsuperscript{6},
  \textbf{Simin Li}\textsuperscript{7},
  \textbf{Xin Yi}\textsuperscript{1,8,$\dagger$},
  \textbf{Hewu Li}\textsuperscript{1}
  \\
  \textsuperscript{1}Tsinghua University, 
  \textsuperscript{2}Tongji University,
  \textsuperscript{3}University of Duisburg-Essen,\\
  \textsuperscript{4}University of Luxembourg,
  \textsuperscript{5}University of Washington,
  \textsuperscript{6}Cornell University,
  \textsuperscript{7}Beihang University,\\
  \textsuperscript{8}Beijing Academy of Artificial Intelligence
  \\
  \texttt{yuwei.chuai@uni.lu},
  \texttt{yixin@tsinghua.edu.cn}
  \\
  \textsuperscript{*}Equal contribution.
  \textsuperscript{$\dagger$}Corresponding authors.
}

\begin{document}
\maketitle
\begin{abstract}
Social media fact-checking has long been challenged by evidence-level and aggregation-level conflicts, where erroneous evidence mimics authoritative news sources. To capture this challenge and support conflict verification tasks, we present \textit{ContraNote}, a large-scale real-world dataset curated from \X's Community Notes system. It includes 33,686 posts for evaluating evidence-level conflict resolution, and 54,474 instances for evaluating aggregation-level prioritization. Additionally, we propose \textit{CoVer}, a factual adjudication framework with three-stage pipelines: evidence schema normalization, factual consensus and support verification. This prioritizes evidence over noise to prevent it from compromising the final verdict. Technical evaluations show that CoVer achieves strong performance compared with state-of-the-art baselines across ContraNote (86.0\% Acc., 68.0\% mac.~F1, 64.5 bal.~Acc. on Conflict; and 88.5\% Acc., 88.5 mac.~F1 and 89.2 bal.~Acc. on Prioritization), CONFACT-HumC (88.4\% Acc.) and CONFACT-ModC (89.4\% Acc.).
\end{abstract}

\section{Introduction}

Developing effective automated fact-checking methods is increasingly important to mitigate the spread of misinformation on social media platforms at scale~\cite{choi2024automated, augenstein2024factuality}. Modern systems commonly adopt a decomposition-aggregation pipeline, which breaks complex claims into atomic sub-claims, verifies each subclaim against external knowledge sources, and synthesizes verdicts to determine overall veracity~\cite{wang2024factcheck}. Retrieval-Augmented Generation (RAG) plays an important role in this pipeline by enabling Large Language Models (LLMs) to ground their reasoning in external evidence retrieved from the open web~\cite{lewis2020retrieval}. 

However, this pipeline frequently encounters conflicts that compromise verdict accuracy. We categorize these conflicts into two distinct levels: (i) \textbf{evidence-level conflict}, where retrieved documents from different sources take opposing stances on the same fact, (ii) \textbf{aggregation-level conflict}, where sub-claims within a single complex claim provide contradictory signals that must be prioritized and reconciled.

\begin{figure}[!htbp]
    \centering 
    \includegraphics[width=0.5\textwidth]{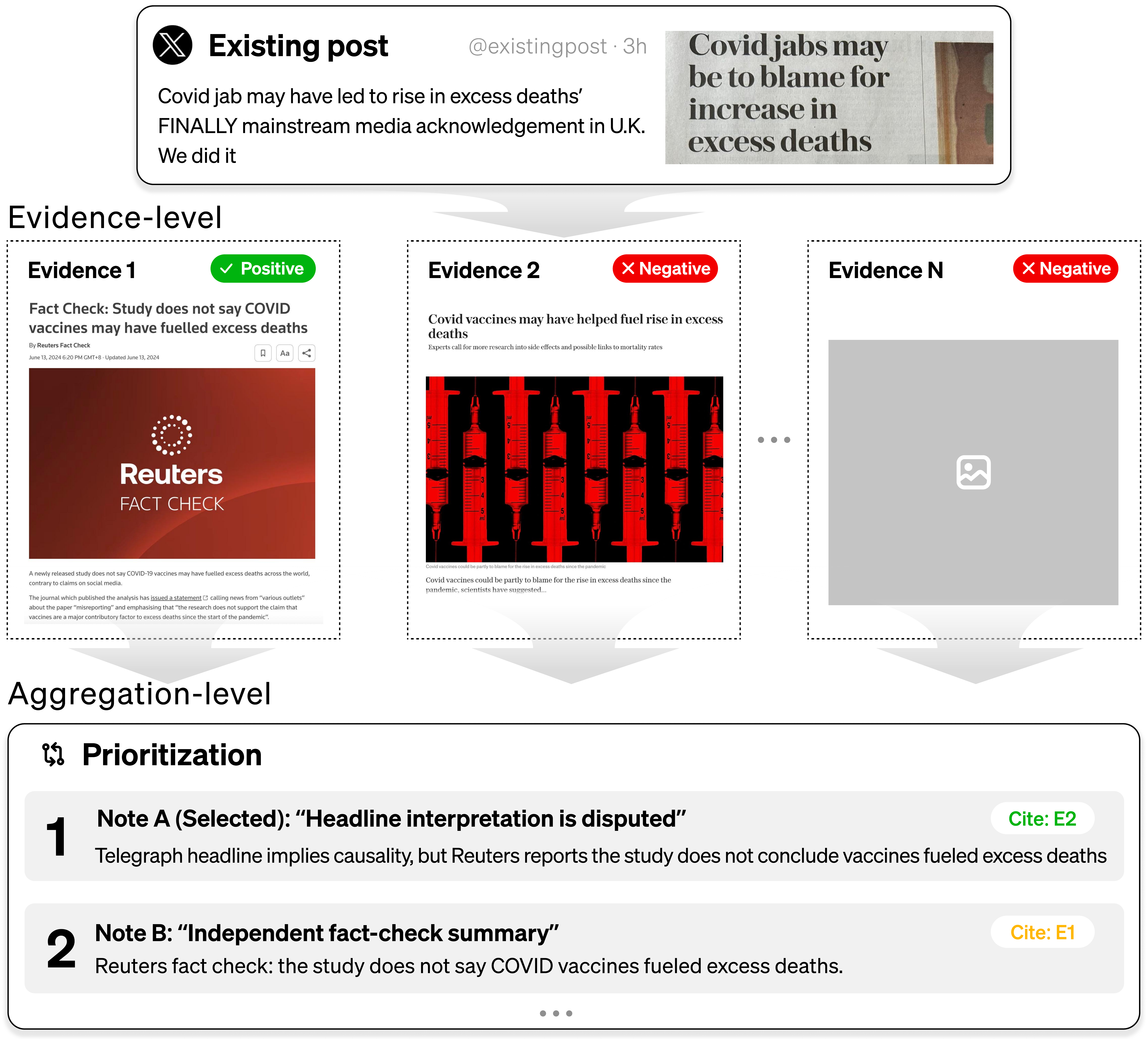}
    \caption{An illustration of conflicting evidence.}
    \label{fig:illustration}
\end{figure}

These challenges are exemplified in a social media claim asserting that a newly published public-health study links COVID-19 vaccines to a rise in excess deaths (Figure~\ref{fig:illustration}). During verification, retrieval may surface an evidence-level conflict: a widely shared headline interprets the study as ``vindicating'' prior anti-vaccine claims, while statements from the publishing journal and epidemiologists clarify that the study has no such causal relationship. Besides, an aggregation-level conflict arises when the claim is decomposed into sub-claims (e.g., trends in excess mortality vs. implied causality), yielding contradictory signals that must be weighed against one another. Here, aggregation-level conflict does not necessarily mean that subclaims contradict one another. Different subclaims may receive local verdicts whose logical implications conflict with the overall claim. For instance, evidence may support the observation that excess mortality increased while refuting the implied causal attribution to COVID-19 vaccination. The conflict therefore arises when local verdicts are aggregated.

Resolving such contradictions is crucial, yet evidence conflict on social media differs from general conflicts in two aspects: (i) \textit{Intentionality:} Unlike general search conflicts that often stem from outdated data, social media conflicts are frequently adversarial, with misinformation crafted to mimic authoritative style. (ii) \textit{Popularity bias:} False narratives on social media may circulate faster than factual corrections. Methods relying on frequency-based aggregation struggle around these issues.

To bridge this gap, we introduce \textit{ContraNote}, a real-world dataset derived from \X's Community Notes system. ContraNote captures real-world conflicts by identifying posts that received opposed debunking statements from crowdsourced contributors. We filtered over two million notes to construct two tasks: a conflict task comprising 33,686 posts to evaluate support/refutation, and a prioritization task comprising 54,474 posts to evaluate the  identification of high-quality evidence. 

We further propose \textit{CoVer}, a framework to parse and adjudicate conflicting evidence by prioritizing evidence over noise. Unlike algorithms that aggregate all evidence at once, CoVer uses a structured pipeline with three modules: evidence schema normalization, factual consensus, and support verification. This facilitates individual scrutinization and filters out noise such as quoted rumors, headlines, or weakly relevant statements.

Technical evaluations show that CoVer outperforms state-of-the-art (SOTA) baselines across various datasets. On ContraNote Conflict and Prioritization, CoVer achieves accuracies of 86.0\% and 88.5\%, with corresponding bal. Acc. of 64.5\% and 89.2\%. It also achieves 88.4\% accuracy on CONFACT-HumC and 89.4\% on -ModC. Ablation studies further confirm the contribution of each module in our proposed framework. Together, this paper makes three contributions:

\noindent $\bullet$ We propose the CoVer framework, an algorithm featuring evidence schema normalization, factual consensus, and support verification modules to effectively resolve evidence conflicts.

\noindent $\bullet$ We construct ContraNote dataset, comprising 33,686 conflicting instances and 54,474 prioritization instances derived from \X, providing testbeds for evaluating real-world conflicts.

\noindent $\bullet$ We provide empirical evidence that CoVer performs strongly relative to SOTA baselines across evidence conflict datasets.

\section{Background and Related Work}

\subsection{Automatic Fact-checking} 

Traditional expert-based fact-checking faces significant challenges regarding scalability, selection bias, and public trust~\cite{pennycook2019fighting, straub2022americans,chuai2025fact,chuai2026community}. In response, community-based and automated alternatives have emerged as viable solutions~\cite{kim2020leveraging, quelle2024perils,zhang2026collab}. Community-based fact-checking, exemplified by \X's Community Notes~\cite{xcommunitynotes}, can achieve accuracy comparable to expert judgments~\cite{allen2021scaling}, resist motivated reasoning~\cite{epstein2020will}, and reach broader online communities~\cite{micallef2020role}. However, it remains too slow to curb misinformation at an early stage and is subject to coordinated rating manipulation~\cite{chuai2024did,chuai2026consensus,chuai2026community}.

Given recent advances in LLMs, automated fact-checking frameworks show promises in identifying suspicious multimodal claims~\cite{qi2024sniffer,zhou2024correcting}, verifying their veracity~\cite{wang2023explainable}, and generating explanations~\cite{yue2024evidence, he2023reinforcement, zeng2024justilm} immediately after publication. Notably, \citet{de2025supernotes} explored synthesizing community notes, while we focus on evidence conflict resolution.

Beyond verification, LLMs can enhance collective decision-making~\cite{yang2024llm} by aggregating diverse perspectives~\cite{burton2024large} and mapping complex opinions to consensus statements~\cite{bakker2022fine}. For instance, \citet{fish2024generative} integrated LLMs with social choice theory to generate multiple summaries than a singular consensus. Our work differs by focusing on evidence conflict resolution.

\subsection{Conflict Resolution in Fact-Checking}

\textbf{Truth discovery.} Early conflict resolution focused on truth discovery, aiming to identify accurate information among conflicting sources by estimating source reliability~\cite{li2016survey}. Traditional methods used iterative probabilistic models to infer trustworthiness~\cite{li2016survey,lyu2017truth}, later evolving into neural frameworks, such as DeClarE, which aggregates external evidence and source credibility via attention mechanisms~\cite{popat2018declare}. Unlike these methods that focus on source credibility, our approach models consensus among conflicting evidence items.

\textbf{Taxonomy and biases in knowledge conflicts.} With the adoption of LLMs, the focus shifted to knowledge conflicts, which can be classified as intra-context, inter-context and parametric discrepancies~\cite{su2024conflictbank,xie2023adaptive,ming2024faitheval}. When resolving these conflicts, LLMs exhibit notable biases: confirmation bias toward internal parametric memory~\cite{xie2023adaptive,su2024conflictbank,ozer2025question}, self-generation bias favoring erroneous self-derived context over retrieved facts~\cite{tan2024blinded}, and selection bias where LLMs detect inconsistencies via Natural Language Inference (NLI)~\cite{jiayang2024econ} but arbitrarily select single evidence items without holistic synthesis~\cite{jiayang2024econ}. To mitigate these biases, we ground conflict resolution in recognizing evidence's stances and resolving based on stance conflicts.

\textbf{Detection and resolution frameworks.} To counter LLM biases, research proposed factual consistency models for detection~\cite{jiayang2024econ}, and employ iterative multi-agent debates (e.g., MADAM-RAG)~\cite{wang2025retrieval}, or contrastive argument synthesis for resolution~\cite{yue2024evidence}. Furthermore, robust resolution requires calibration and uncertainty estimation to merge conflicting and evolving evidence in temporal contexts~\cite{wan2024evidence,chen2022rich,ozer2025question,burton2024large}. Unlike frameworks designed for document retrieval and verification, we focused on reconciling contradictory evidence.

\section{Problem Definition}\label{sec:problem_definition}

In automated fact-checking, conflicting evidence primarily exist at two stages: the \textit{evidence} level and the \textit{aggregation} level. \textit{Evidence}-level conflict occurs when retrieved documents present contradictory stances on a single fact. \textit{Aggregation}-level conflict arises when a complex claim is decomposed into sub-claims that yield divergent verdicts. For example, a correct attribution alongside false causality requires the system to synthesize these mixed signals into a coherent conclusion.

Formally, let a claim $C$ be decomposed into subclaims $S = \{s_1, ..., s_n\}$. For each $s_i$, the system retrieves a set of evidence documents $E_i = \{e_{i, 1}, ..., e_{i,m}\}$. An evidence-level conflict exists if $E_i$ contains contradictory stance labels that simultaneously support and refute $s_i$. An aggregation-level conflict occurs when the set of local verdicts $V_S = \{v(s_1), ..., v(s_n)\}$ has opposing logical implications for the overall veracity of $C$. The objective is to learn a verification function $\mathcal{F} (C, \bigcup E_i) \rightarrow y$ that maps the claim and conflicting evidence to a final verdict $y \in \{\text{Supported}, \text{Refuted}, \text{Not}\,\text{Enough}\,\text{Information}$ by prioritizing evidence over noise.

\section{ContraNote Dataset}\label{sec:dataset}
 
We constructed the ContraNote dataset using the open-source Community Notes repository published by \X~\cite{communitynotesdata}. Community Notes is a crowd-sourced misinformation debunking mechanism where qualified contributors provide additional context to evaluate the veracity of posts. The longitudinal data used in this study span from June 2021 to May 2026. 

\begin{figure*}[!htbp]
    \includegraphics[width=\textwidth]{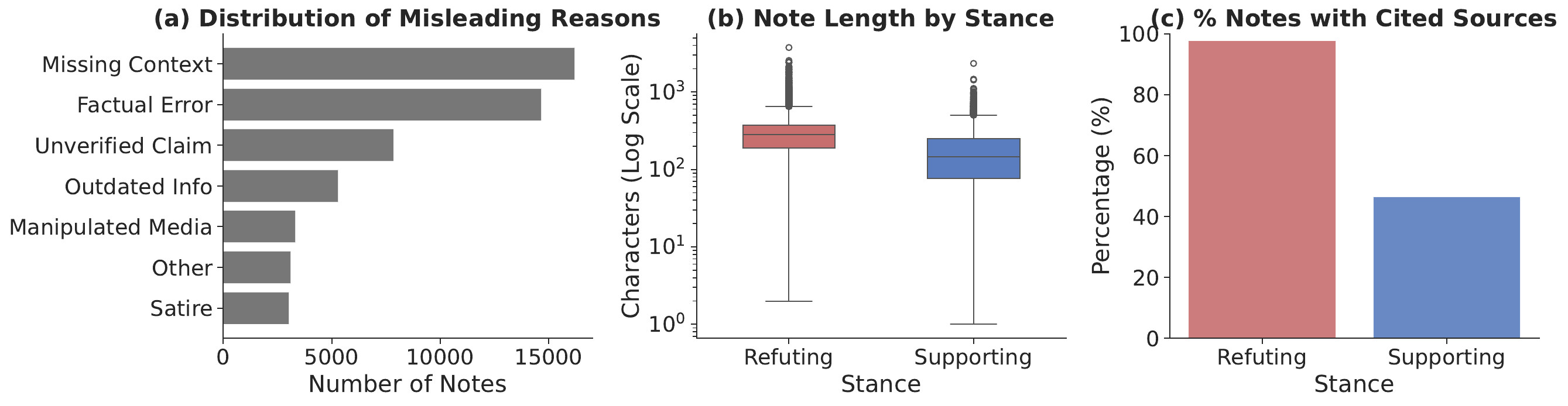}
    \caption{Distributions in ContraNote, (a) distribution of misleading reasons, (b) note length by stance, (c) percentage of notes with cited sources.}
    \label{fig:comnote}
\end{figure*}

Individual posts on the platform frequently elicit multiple community notes with divergent viewpoints. Some contributors may flag a post as potentially misleading, while others may argue that it is not misleading because it is factually correct, satirical, or containing personal opinion. These conflicting perspectives on a single post create a complex environment of evidentiary contradictions. Final note selection is determined by user helpfulness ratings and algorithmic prioritization. This inherent complexity provides a unique opportunity for LLMs to learn from real-world conflicts and develop mechanisms for information prioritization.

Our initial corpus comprised 2,276,724 community notes corresponding to 1,502,486 unique posts. To isolate instances of conflict, we first identified 443,148 posts that received more than one community note. We then focused on posts containing stance contrast, defined by the two Community Notes classifications: \nolinkurl{MISINFORMED\_OR\_POTENTIALLY\_MISLEADING} and \nolinkurl{NOT\_MISLEADING}.

We constructed two benchmark tasks from this filtered corpus. The first, \textbf{ContraNote Conflict}, evaluates whether an original post should be supported or refuted given conflicting notes. Refuted instances are posts satisfying three conditions: (1) the notes have stance contrast, (2) it has at least one note rated as \nolinkurl{CURRENTLY\_RATED\_HELPFUL}, and (3) at least one helpful note is labeled \nolinkurl{MISINFORMED\_OR\_POTENTIALLY\_MISLEADING}. These instances correspond to claims for which the crowd-rated consensus supports active correction or refutation. This yields 27,445 \textit{Refuted} instances. Supported instances should satisfy: (1) the notes have stance contrast, (2) it contain no helpful misleading note, and all misleading notes are rated \nolinkurl{CURRENTLY\_RATED\_NOT\_HELPFUL}, (3) the majority of its notes are labeled \nolinkurl{NOT\_MISLEADING}. We apply different criteria as our dataset from Community Notes have no non-misleading notes with helpful status. This process yields 6,241 \textit{Supported} instances.  \textbf{ContraNote Conflict} contains 33,686 posts, including 27,445 \textit{Refuted} and 6,241 \textit{Supported} instances.

The second task, \textbf{ContraNote prioritization}, evaluates whether a model can identify potentially conflicting high-quality evidence among notes. We selected posts with stance contrast that contain at least one helpful note and one non-helpful note, where non-helpful candidates include notes rated as \nolinkurl{CURRENTLY\_RATED\_NOT\_HELPFUL} or \nolinkurl{NEEDS\_MORE\_RATINGS}. Given the post context and its set of notes, the model needs to predict which note should be prioritized. This produces a balanced benchmark of 54,474 instances over 27,237 posts, with 27,237 \textit{Supported} target notes and 27,237 \textit{Refuted} target notes. 

As shown in Figure~\ref{fig:comnote}, missing context (22,615 instances) and factual errors (20,918 instances) are primary drivers of misleading claim. Refuting notes exhibit greater detail (1.77 per post) and length (289.08 characters) compared to supporting ones. Evidence link analysis reveals that news (49.1\%) and social media platforms (24.6\%) are major cited sources, while encyclopedic references remain secondary. This indicates that ContraNote primarily relies on heterogeneous evidence that require provenance assessment. Qualitative coding reveals that relations between contradictory notes extend beyond direct contrasts, which include contextual, scoping and aggregation-level disagreements. Notably, \textit{note-necessity disagreement} is the primary conflict pattern (33\%), where contributors contest the necessity of moderation. Language distribution analysis shows that English constitutes the majority language (13,776, 62.98\%), followed by Spanish (1,758, 8.04\%) and Portuguese (1,368, 6.25\%). 

As shown in Table~\ref{tab:dataset_positioning}, ContraNote extends prior benchmarks~\cite{augenstein2019multifc,schlichtkrull2023averitec,chen2022generating} by capturing naturally occurring evidence conflicts within single posts. To evaluate label reliability, three trained annotators independently annotated on randomly sampled 500 instances, which yield high inter-rater reliability (Fleiss' $\kappa=$0.82). Majority-vote human annotations aligned with dataset labels in 94.2\% cases, validting labels' accuracy. To account for potential bias, we analyzed using PoliticalBiasBERT, and found the dataset covered broad political orientations (18.6\% left, 47.0\% center, 34.4\% right) and topic domains (e.g., 26.0\% political/governance, 20.6\% science/technology, 29.8\% media/entertainment/sports). This indicates that ContraNote reflects the consensus signal generated by Community Notes mechanism. Details are all shown in Appendix~\ref{sec:dataset_analysis}.

\section{CoVer}\label{sec:technique} 


\subsection{Algorithm Overview}

\begin{figure*}[!htbp]
    \centering
    \includegraphics[width=0.86\textwidth]{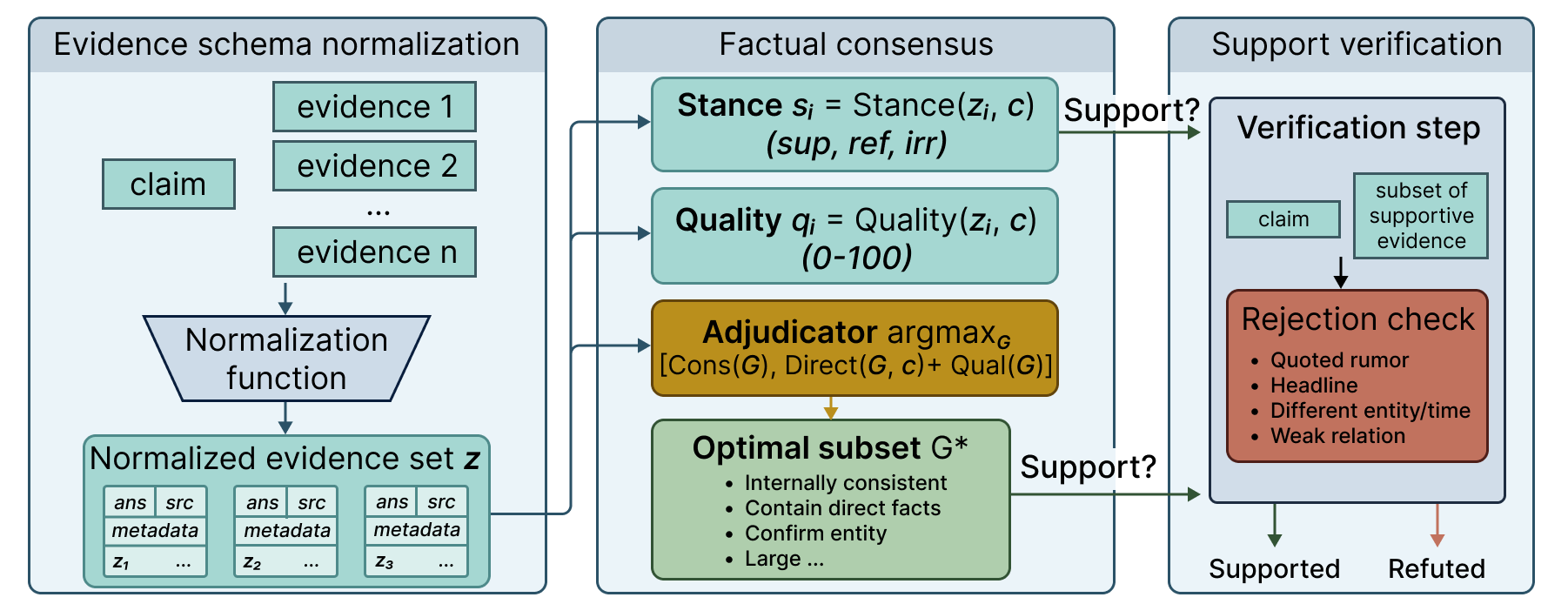}
    \caption{The CoVer framework.}
    \label{fig:framework}
\end{figure*}

As shown in Figure~\ref{fig:framework}, given a claim $q$ and an evidence set $\mathcal{E}=\{e_i\}_{i=1}^{N}$, CoVer predicts a label $y \in \{\text{Supported}, \text{Refuted}\}$. This algorithm has three modules: evidence schema normalization, factual consensus, and support verification.

\subsection{Evidence Schema Normalization}

Each evidence item $e_i$ may contain free text and structured metadata $e_i = (t_i, m_i)$, where $t_i$ is textual content and $m_i$ contains optional fields such as source URL. We normalize each item into a canonical representation: $z_i = \phi(e_i) = \phi(t_i, m_i).$ The normalized item $z_i$ preserves both text and schema-level cues. For QA evidence, we retained proposed answer fields. For source-pointer evidence, we retained page and line identifiers. For candidate-note evidence, status, label, stance, and helpfulness fields are kept. For plain fact-checking evidence, article text and snippets are kept. This gives a unified evidence set $\mathcal{Z}=\{z_i\}_{i=1}^{N}$.

\subsection{Factual Consensus}

The second stage performs factual adjudication over the normalized evidence via a three-step pipeline: individual evidence adjudication, consensus aggregation, and final verdict generation.

For each normalized evidence item $z_i \in \mathcal{Z}$ associated with the claim $q$, CoVer estimates its stance toward the claim and evaluates its evidential quality. Instead of using separate prompts, we use a single LLM call with structured output constraints to jointly predict the stance $s_i$ and four fine-grained quality components. These constraints require a valid stance from the predefined label set, numeric quality scores in $[0,1]$, and the presence of all fields needed by deterministic aggregation: 

\begin{equation}
(s_i, d_i, a_i, l_i, u_i) = \mathrm{LLM}_{\mathrm{adj}}(q, z_i, m_i),
\end{equation}
where $s_i \in \{\mathrm{support}, \mathrm{refute}, \mathrm{irrelevant}\}$, and $m_i$ denotes the preserved metadata schema (e.g., source URL, helpfulness signals). The quality components are scored on a normalized scale $[0,1]$ according to the following rubrics:

\noindent $\bullet$ \textit{Directness} ($d_i$) measures whether $z_i$ directly addresses the central proposition of $q$, penalizing items that merely share superficial entities.

\noindent $\bullet$ \textit{Attribute alignment} ($a_i$) evaluates factual compatibility across key dimensions, including entity identity, temporal scope, location, and numerical arguments.

\noindent $\bullet$ \textit{Schema reliability} ($l_i$) includes metadata cues $m_i$ (e.g., source authority and community helpfulness ratings) to assess the trustworthiness of the evidence channel.

\noindent $\bullet$ \textit{Informativeness} ($u_i$) quantifies the substantive factual content, assigning low scores to repetitive rumor quotes, headlines, or text that merely reports the existence of a claim.

The overall quality score $q_i$ is deterministically computed as a weighted linear combination of these components: $q_i = \lambda_d d_i + \lambda_a a_i + \lambda_l l_i + \lambda_u u_i$, where weights are tuned as $\lambda_d = \lambda_a = \lambda_l = \lambda_u = 0.25$ to balance each dimension (see Appendix~\ref{app:hyperparameters}). Given $(s_i, q_i)$ for all evidence items, CoVer aggregates the evidence deterministically to resolve conflicts. We first compute quality-weighted cumulative scores for the supporting and refuting stances:

\begin{equation}
A_{\mathrm{sup}} = \sum_{i=1}^{|\mathcal{Z}|} q_i \cdot \mathbb{I}[s_i = \mathrm{support}], 
\end{equation}
\begin{equation}
A_{\mathrm{ref}} = \sum_{i=1}^{|\mathcal{Z}|} q_i \cdot \mathbb{I}[s_i = \mathrm{refute}],
\end{equation}

where $\mathbb{I}[\cdot]$ is the indicator function. The dominant factual stance $s^*$ is selected by comparing the cumulative strengths:

\begin{equation}
s^* = \begin{cases}
\mathrm{support}, & \text{if } A_{\mathrm{sup}} > A_{\mathrm{ref}}, \\
\mathrm{refute}, & \text{if } A_{\mathrm{ref}} \ge A_{\mathrm{sup}}.
\end{cases}
\label{eq:stance_tiebreak}
\end{equation}

We use Refute as the tie-breaker, which follows the conservative goal of avoiding unsupported positive predictions. To assess potential biases, we evaluate a variant in which ties are assigned to Support. Corresponding results are reported in Appendix~\ref{app:tiebreaking}. To remove irrelevant noise and low-quality assertions, the final consensus evidence set $G^*$ is filtered using a quality threshold $\tau_q$:

\begin{equation}
G^* = \{z_i \in \mathcal{Z} \mid s_i = s^* \land q_i \ge \tau_q\}.
\end{equation}

The factual correlation between the aggregated consensus set $G^*$ and claim $q$ is adjudicated by an additional LLM call, yielding $r = \mathrm{LLM}_{\mathrm{ver}}(q, G^*) \in \{\mathrm{entails}, \mathrm{contradicts}, \mathrm{insufficient}\}$. $\mathrm{entails}$ is mapped to Supported, while others are mapped to Refuted, as neither outcome suggests that the original post is supported under the Community Notes labeling protocol. To test the effect of this strategy, we evaluate a Supported/Partially Supported/Refuted setting in Appendix~\ref{app:three_way}.

\subsection{Support Verification}

Given the factual consensus output $y_{\mathrm{strict}}$, CoVer applies support verification only when factual consensus predicts Supported. If $y_{\mathrm{strict}}=\mathrm{Refuted}$, the algorithm terminates and returns Refuted. We use this as a conservative filter for positive predictions.
When $y_{\mathrm{strict}}=\mathrm{Supported}$, CoVer constructs the candidate support set
\[
\mathcal{Z}_{\mathrm{sup}} =
\{z_i \in \mathcal{Z}: s_i=\mathrm{support},\ q_i \ge \tau_q\}.
\]
It then performs a final verification call: $v=g_\theta(q,\mathcal{Z}_{\mathrm{sup}})
\in\{\mathrm{valid},\mathrm{invalid}\}.$
This verifier checks whether the selected supporting evidence directly and independently validates the claim's central proposition. It rejects support if the evidence merely quotes a claim or rumor, a headline or fact-check setup, about a different entity, time, answer, or scope, or is merely related without factual statement. The final prediction $\hat{y}$ is labeled as $\mathrm{Supported}$ if $y_{\mathrm{strict}} = \mathrm{Supported} \land v = \mathrm{valid}$, and $\mathrm{Refuted}$ otherwise.

\section{Experiments}\label{sec:experiments}

\subsection{Datasets} We choose various datasets representing different conflict levels. Specifically, we examine conflicting evidence in social media fact-checking and other scenarios to test CoVer's generalizability:

\noindent \textbf{CONFACT~\cite{ge2025resolving}.} Unlike traditional benchmarks where evidence is often consistent, CONFACT is specifically curated to include claims with opposed evidence on the web (e.g., conflicting reports on political events or scientific debates). It serves as the primary testbed for measuring agents' ability to resolve evidence conflicts.

\noindent \textbf{ConflictBank~\cite{su2024conflictbank}.} This benchmark analyzes model behavior by simulating knowledge conflicts. It includes 553,117 QA pairs derived from 2,863,205 Wikidata claims, covering three main conflict causes: misinformation, temporal change, and semantic variation. Using the original QA pairs, we construct refutation examples by treating the modified evidence as conflicting evidence groups, yielding 1,659,351 data items.

\noindent \textbf{ECON~\cite{jiayang2024econ}.} The dataset is based on two public datasets: Natural Questions and Complex Web Questions, where they constructed alternative answers as conflicting evidence, producing different types of answer and factoid conflicts: degree, entity, negation, number, temporal, verb, and other types. ECON contains 4,995 data items.

\noindent \textbf{ContraNote.} We use the version described in Sec.~\ref{sec:dataset}. 
The primary binary labels follow the Community Notes labeling scheme. We also provide a three-way pilot analysis in Appendix~\ref{app:three_way}.

\noindent \textbf{FEVER~\cite{thorne2018fever}.} It is a most widely used fact-check dataset~\cite{min2023factscore,zhao2023felm}, featuring fact extraction and verification. It contains claims generated by altering sentences extracted from Wikipedia and subsequently verified without access to the source sentences. Claims are labeled supported, refuted and not enough information. We used the shared claim subset, containing 19,998 claims.

\begin{table}[ht]
\centering
\resizebox{0.5\textwidth}{!}{
\begin{tabular}{llccc}
\toprule
\textbf{Dataset} & \textbf{Subset} & \textbf{Number} & \textbf{Positive} & \textbf{Negative} \\ \midrule
\multirow{2}{*}{CONFACT} & HumC & 287 & 51 & 236 \\
 & ModC & 611 & 125 & 486\\ 
ConflictBank & -- & 1,659,351 & 553,117 & 1,106,234 \\
ECON & -- & 4,995 & 2,043 & 2,952 \\
\multirow{2}{*}{ContraNote} & Conflict & 33,686 & 6,241 & 27,445 \\
& Prioritization & 54,474 & 27,237 & 27,237 \\
FEVER & -- & 13,332 & 6,666 & 6,666 \\ \bottomrule
\end{tabular}
}%
\caption{Dataset distribution (Positive: Support, Negative: Refute).}
\label{tab:dataset_statistics}
\end{table}

\subsection{Baselines} 

We compare CoVer with eight representative baselines for conflict resolution or social media fact-checking. To ensure a controlled evaluation, we decouple verification from retrieval. By providing all methods with identical sets of conflicting evidence, we isolate retrieval variance as a confounding variable. Therefore, while some baselines originally included retrieval components, we adapt them to focus on evidence adjudication. 

\noindent \textbf{FacTool~\cite{chern2023factool}:} A method that performs a single-pass verification based on retrieved evidence, representing a basic fact-check flow.

\noindent \textbf{FactCheckGPT~\cite{wang2024factcheck}:} A decomposition-based method, which breaks a claim into atomic sub-claims, retrieves evidence for each sub-claim individually, and aggregates the results, thereby serving a standard automated fact-checking pipeline.

\noindent \textbf{FIRE~\cite{xie2025fire}:} An iterative reasoning agent. Unlike FacTool, FIRE operates as a loop. It assesses whether the current context is sufficient to answer the claim. If not, it considers new evidence. Through this process, it implicitly models conflict.

\noindent \textbf{Confact~\cite{ge2025resolving}:} A source-aware RAG framework designed to resolve evidentiary conflicts by integrating media background metadata (e.g., source credibility ratings and bias information) directly into the answer generation stage. It uses structured reasoning (e.g., Chain-of-Thought) to evaluate and prioritize evidence from trustworthy sources, thereby mitigating the influence of misleading information from unreliable origins.

\noindent \textbf{ECON~\cite{jiayang2024econ} (i.e., ConflictRes):} A framework focusing on evidence conflicts, especially those occurring between different retrieved context. ECON addresses the gap between LLMs' detection and their unreliable resolution behaviors, such as arbitrary evidence selection or over-reliance on internal priors.

\noindent \textbf{Additional baselines:} We also compare an AVeriTeC-style verifier, MADAM-RAG, and ClaimDecomp. The AVeriTeC-style verifier uses question-guided evidence verification for real-world claims, while MADAM-RAG represents a multi-agent retrieval-augmented verification pipeline. ClaimDecomp decomposes a complex claim into literal and implied subclaims, verifies them individually, and aggregates their local verdicts. Under our fixed-evidence setting, these methods receive the same claim and evidence pool as the other baselines, isolating differences in verification and aggregation rather than retrieval coverage.

\subsection{Study Settings} 

\noindent \textbf{Implementation details}\\
All agents in our evaluation use \texttt{gpt-4o} as the backbone LLMs. Baselines are implemented with the following configurations to ensure reproducibility: \textbf{FacTool} generates two search queries and retrieves the top-10 results per query using a CoT verification process. \textbf{FactCheckGPT} decomposes claims into 2–3 queries and verifies results through NLI. The iterative \textbf{FIRE} agent is restricted to a maximum of 10 steps. \textbf{CONFACT} retrieves the top-10 results and augments them with claim background descriptions of under 50 words each, while \textbf{ConflictRes} focuses on resolving discrepancies between the top-10 retrieved snippets. Similarly, we limit CoVer to a maximum of 10 evidence items to maintain processing efficiency. Each method receives the same query and evidence. We repeat each experiment five times and report the average results.

\vspace{.5em}
\noindent \textbf{Ablation Settings}\\
We evaluate the contribution of each CoVer component by removing modules. 

\noindent \textbf{Without evidence \textit{schema} normalization:} we remove the schema normalization module and provide evidence to the adjudicator only as unstructured text. We omit structured cues such as proposed answers, source pointers, line identifiers, candidate-note stances, and target-candidate markers. This tests whether task-relevant evidence structures are necessary for conflict resolution.

\noindent \textbf{Without factual \textit{consensus}:} the system no longer selects direct, internally consistent, and claim-aligned evidence group before making a decision. This evaluates whether factual consensus with evidence group is helpful.

\noindent \textbf{Without \textit{support} verification:} the model accepts support decisions without additional check for direct entailment, target-candidate validity, or contradiction by stronger evidence. 

\noindent \textbf{Pairwise removals:} we further evaluate all pairwise removals: w/o Schema + Consensus, w/o Schema + Support, and w/o Consensus + Support. These settings measure whether the modules provide complementary benefits or whether performance is driven by a single component.

\noindent \textbf{Without all:} we remove all three modules to create the minimal setting.

\subsection{Main Results}\label{sec:main_results}

Table~\ref{tab:baseline_comparison} compares CoVer with the baselines across all datasets.

\begin{table*}[t]
\centering
\small
\resizebox{\textwidth}{!}{
\begin{tabular}{lccccccc}
\toprule
\textbf{Method} & \makecell[c]{\textbf{CONFACT}\\\textbf{HumC}} & \makecell[c]{\textbf{CONFACT}\\\textbf{ModC}} & \textbf{ConflictBank} & \textbf{ECON} & \textbf{FEVER} & \makecell[c]{\textbf{ContraNote}\\\textbf{Conflict}} & \makecell[c]{\textbf{ContraNote}\\\textbf{Prioritization}} \\
\midrule
FacTool & \res{80.0}{60.8}{59.9} & \res{81.5}{69.6}{68.2} & \res{57.8}{52.0}{52.5} & \res{70.0}{69.2}{69.2} & \res{51.0}{50.3}{61.7} & \res{71.1}{63.4}{71.2} & \res{59.8}{59.8}{60.0} \\
FactCheckGPT & \res{84.0}{67.7}{65.8} & \res{80.5}{68.0}{66.7} & \res{63.5}{57.5}{58.0} & \res{67.5}{66.6}{66.6} & \res{81.5}{80.9}{85.0} & \res{72.2}{60.1}{62.9} & \res{55.8}{55.7}{56.3} \\
FIRE & \res{79.0}{55.0}{54.6} & \res{78.4}{66.0}{65.4} & \res{54.8}{48.2}{48.3} & \res{63.0}{61.9}{62.0} & \res{72.0}{71.7}{76.7} & \res{72.2}{63.0}{68.5} & \res{64.5}{63.8}{63.9} \\
Confact & \res{75.9}{62.5}{64.4} & \res{80.8}{74.2}{77.4} & \res{60.5}{56.4}{57.9} & \res{74.0}{74.0}{74.4} & \res{87.0}{85.1}{84.3} & \res{82.8}{73.4}{76.1} & \res{62.8}{62.8}{62.9} \\
ConflictRes & \res{81.5}{64.2}{63.1} & \res{77.5}{68.5}{70.0} & \res{57.0}{44.1}{44.7} & \res{68.5}{67.8}{67.8} & \res{74.5}{73.4}{76.0} & \res{81.3}{70.2}{71.8} & \res{61.5}{60.1}{60.6} \\
CoVer & \res{88.4}{77.1}{74.3} & \res{89.4}{83.1}{81.6} & \res{73.5}{63.4}{62.5} & \res{77.5}{77.5}{77.7} & \res{93.4}{92.9}{94.3} & \res{86.0}{68.0}{64.5} & \res{88.5}{88.5}{89.2} \\
\bottomrule
\end{tabular}
}
\caption{Baseline comparison across various evaluation datasets. Metrics in each cell are formatted as Accuracy $\mid$ \textcolor{blue}{mac. F1} $\mid$ \textcolor{purple}{bal. Acc.}.}
\label{tab:baseline_comparison}
\end{table*}

\begin{table*}[t]
\centering
\small
\resizebox{\textwidth}{!}{
\begin{tabular}{lcccccccc}
\toprule
\textbf{Setting} & \makecell[c]{\textbf{CONFACT}\\\textbf{HumC}} & \makecell[c]{\textbf{CONFACT}\\\textbf{ModC}} & \textbf{ConflictBank} & \textbf{ECON} & \textbf{FEVER} & \makecell[c]{\textbf{ContraNote}\\\textbf{Conflict}} & \makecell[c]{\textbf{ContraNote}\\\textbf{Prioritization}} \\
\midrule
Full & \res{88.4}{77.1}{74.3} & \res{89.4}{83.1}{81.6} & \res{73.5}{63.4}{62.5} & \res{77.5}{77.5}{77.7} & \res{93.4}{92.9}{94.3} & \res{86.0}{68.0}{64.5} & \res{88.5}{88.5}{89.2} \\
w/o Schema & \res{88.4}{77.1}{74.3} & \res{89.4}{83.1}{81.6} & \res{68.8}{56.8}{56.7} & \res{45.5}{40.7}{47.0} & \res{82.4}{81.6}{84.5} & \res{84.0}{71.7}{71.2} & \res{76.9}{76.9}{77.0} \\
w/o Consensus & \res{84.4}{74.0}{75.4} & \res{81.1}{73.9}{76.4} & \res{63.3}{60.8}{64.0} & \res{77.5}{77.5}{77.7} & \res{88.5}{87.6}{89.5} & \res{69.0}{58.0}{49.7} & \res{42.5}{31.9}{40.2} \\
w/o Support & \res{85.8}{73.2}{71.6} & \res{88.4}{81.5}{80.1} & \res{70.0}{60.0}{59.5} & \res{77.5}{77.5}{77.7} & \res{88.5}{87.6}{89.5} & \res{85.5}{66.3}{63.1} & \res{68.5}{68.3}{68.3} \\
w/o Schema + Consensus & \res{80.6}{67.0}{67.4} & \res{78.9}{70.3}{72.3} & \res{63.8}{61.3}{64.3} & \res{76.5}{76.5}{77.0} & \res{88.0}{86.8}{87.6} & \res{60.0}{55.7}{65.6} & \res{54.0}{59.8}{53.5} \\
w/o Schema + Support & \res{82.7}{62.3}{60.5} & \res{86.3}{76.3}{73.4} & \res{69.3}{60.0}{59.6} & \res{59.5}{55.0}{57.3} & \res{86.5}{85.5}{87.3} & \res{84.0}{71.7}{71.2} & \res{79.4}{79.4}{79.5} \\
w/o Consensus + Support & \res{84.4}{74.0}{75.4} & \res{81.1}{73.9}{76.4} & \res{63.3}{60.8}{64.0} & \res{77.5}{77.5}{77.7} & \res{88.5}{87.6}{89.5} & \res{72.0}{60.9}{52.6} & \res{40.5}{30.1}{38.3} \\
w/o All & \res{80.6}{67.0}{67.4} & \res{78.9}{70.3}{72.3} & \res{63.8}{61.3}{64.3} & \res{76.5}{76.5}{77.0} & \res{88.0}{86.8}{87.6} & \res{61.0}{56.4}{67.4} & \res{49.7}{54.8}{49.3} \\
\bottomrule
\end{tabular}
}
\caption{Ablation results on different components of CoVer, formatted as Accuracy $\mid$ \textcolor{blue}{mac. F1} $\mid$ \textcolor{purple}{bal. Acc.}. Schema=Evidence Schema Normalization, Consensus=Factual Consensus, Support=Support Verification.}
\label{tab:full_ablation}
\end{table*}

\textbf{\textit{CoVer performs strongly on tasks involving complex contradictions and ambiguity.}} As shown in Table~\ref{tab:baseline_comparison}, CoVer surpasses all baselines on CONFACT, achieving 88.4\% accuracy on HumC and 89.4\% on ModC. On ContraNote Conflict, it attains a leading accuracy of 86.0\%, compared with 82.8\% for Confact and 81.3\% for ConflictRes. These results show CoVer's ability to synthesize conflicting information and adjudicate claims involving nuanced inconsistencies.

\textbf{\textit{CoVer is effective at evidence prioritization and domain-specific conflict resolution.}} On ContraNote Prioritization, CoVer achieves 88.5\% accuracy, exceeding baselines such as FIRE (64.5\%). Similarly, on ConflictBank, CoVer achieves 73.5\% accuracy, showing marked improvement over FactCheckGPT (63.5\%). This highlights CoVer's capacity to process structured conflicting scenarios and prioritize reliable signals.

\textbf{\textit{Beyond conflict arbitration, CoVer exhibits high accuracy on fact-check datasets.}} It achieved 93.4\% accuracy on FEVER, surpassing Confact (87.0\%) and FactCheckGPT (81.5\%). On ECON, it achieves the highest accuracy (77.5\%) versus Confact (74.0\%). This confirms CoVer's generalizability to fact-check datasets.

\vspace{.5em}
\noindent \textbf{Additional baseline comparisons:}
The AVeriTeC-style verifier obtains 65.6 and 80.5 mac. F1 on ContraNote Conflict and Prioritization, respectively, while MADAM-RAG obtains 47.7 and 70.3; CoVer obtains 68.0 and 88.5 on the same two tasks. ClaimDecomp obtains accuracies of 82.23, 84.58, 80.00, and 66.00, with corresponding mac. F1 scores of 67.28, 76.21, 54.29, and 65.58 on CONFACT-HumC, CONFACT-ModC, ContraNote Conflict, and ContraNote Prioritization, respectively. Under the same dataset order, CoVer obtains mac. F1 scores of 77.10, 83.10, 68.00, and 88.50. These results show that CoVer remains competitive with retrieval-oriented and claim-decomposition baselines, with the largest gains appearing on evidence prioritization and conflict aggregation.

\vspace{.5em}
\noindent \textbf{Retrieval:}
Beyond gold evidence setting, which isolates evidence adjudication and prevents confounding effects from retrieval quality, we assess whether the framework remains useful with end-to-end evidence retrieval. We pair each method with upstream retriever and evaluate the resulting claim-level predictions. On retrieval-enabled ContraNote Conflict setting, CoVer obtains 69.6 mac. F1, compared with 59.2 mac. F1 for strongest baseline. These suggests that CoVer complements retrieval, adjudicating evidence with varied stances, reliability and claim alignment. 

\vspace{.5em}
\noindent \textbf{Shortcut baseline}\\
To test whether ContraNote labels are recoverable from inputted metadata, we evaluate a shortcut baseline, with rules that predict Refuted if and only if at least one note is marked both helpful and misleading. On ContraNote Conflict, this obtains 81.5\% accuracy, 44.9 mac. F1 and 50.0 bal. Acc. On ContraNote Prioritization, this obtains 50.0\% accuracy, 33.3 mac. F1 and 50.0 bal. Acc. Its high accuracy is explained by class imbalance, where mac. F1 and bal. Acc. are by chance.

\vspace{.5em}
\noindent \textbf{Metadata-suppressed evaluation}\\
We further test conditions of CoVer by removing note-status fields that could expose construction-time signals, i.e., target note's helpfulness, status, classification, and label fields. In this setting, CoVer obtains 87.5\% mac. F1 and 86.5\% bal. Acc. This shows that CoVer retains strong performance without access to metadata fields.

\vspace{.5em}
\noindent \textbf{Statistical testing}\\
We assess pairwise differences using McNemar's test with $\alpha=0.05$. CoVer's improvements are significant on all evaluated datasets except for comparisons on ContraNote Conflict with CONFACT and ConflictRes. Accordingly, the results on ContraNote Conflict indicate a positive performance trend but are not significant. Appendix~\ref{app:error_analysis} complements these tests with error analysis.

\subsection{Ablation Study}\label{sec:ablation}

\textbf{\textit{Factual consensus module is critical for resolving complex contradictions.}} As shown in Table~\ref{tab:full_ablation}, removing this module (w/o Consensus) degrades performance in tasks requiring nuanced arbitration, where accuracy on ContraNote Prioritization drops from 88.5\% to 42.5\%. Similarly, accuracy on ContraNote Conflict drops from 86.0\% to 69.0\%.

\textbf{\textit{Evidence schema normalization is essential for parsing factual information.}} While removing this module leaves performance on CONFACT unaffected, it causes degradation on ECON, failing from 77.5\% to 45.5\%. We also observe substantial degradations on FEVER (93.4\% to 82.4\%) and ConflictBank (73.5\% to 68.8\%). 

\textbf{Support verification ensures reasoning stability, and CoVer exhibits strong synergistic effects when integrating all modules.} Removing support verification degrades performance across multiple datasets, most notably on ContraNote Prioritization (dropping from 88.5\% to 68.5\%). Furthermore, w/o all configuration produces the most substantial degradation on complex tasks, reducing CONFACT-HumC accuracy to 80.6\% and ContraNote Prioritization to 49.7\%. 

\subsection{Temporal and Paraphrase Robustness}

As GPT-4o may be trained on publicly available Community Notes, we evaluate whether CoVer relies on memorization. We construct a temporally held-out slice from January 2026 and paraphrase the claims while preserving their semantics. On this test, CoVer obtains 68.8 mac. F1, compared with 66.5 for the strongest baseline. 

\subsection{Computational Cost Analysis}

Based on results from all evaluation datasets, the average per-call generation, and end-to-end fact-checking time are 5.8s. Average token usage is 1546.1 tokens per request (1274.9 input / 271.2 output), corresponding to an estimated cost of \$0.0059 per request. These results suggest that resolving conflicting evidence with CoVer remains computationally and economically feasible. To control for inference budget, we evaluate both single-call baseline configurations and multi-call configurations matched to the number of LLM calls used by CoVer. Multi-call versions improve some baselines, but CoVer remains competitive under matched call budgets. Call counts and token usage are reported in Appendix~\ref{app:implementation}.

\section{Conclusion}

This paper addresses evidence-level and aggregation-level conflicts in automated social media fact-checking. We propose CoVer, a framework that resolves these contradictions through structuring evidence schema normalization, factual consensus, and support verification, effectively prioritizing evidence over noise. We then construct ContraNote, a real-world dataset derived from \X for conflict resolution (33,686 items) and evidence prioritization (54,474 items). Extensive experiments show that CoVer achieved strong performance compared with SOTA baselines, achieving accuracies of 86.0\% and 88.5\% on the Conflict and Prioritization tasks (bal. Acc.: 64.5\% and 89.2\%) respectively.

\section*{Acknowledgments}

This work was supported by Beijing Major Science and Technology Project under Contract no. Z251100008125024, Beijing Academy of Artificial Intelligence (BAAI), and the Luxembourg National Research Fund (ref. C25/IS-SAS/19599536).

\section{Limitations}

We acknowledge several limitations in this paper that highlight directions for future research.

First, the proposed framework and the ContraNote dataset focus exclusively on textual claims and metadata, predominantly in English. Although we diversify the language coverage, the current coverage remains insufficient for comprehensive real-world deployment. Furthermore, modern social media misinformation is multimodal. Our current setting excludes conflict adjudication involving manipulated images, deepfakes, or out-of-context videos, which frequently drive real-world evidence contradictions. Additionally, the efficacy of the framework in low-resource languages or highly specialized domains (e.g., legal or medical texts) requires further validation.

Second, the ground-truth definition in the ContraNote dataset relies on crowdsourced consensus and algorithmic helpfulness scores. While this approach reflects practical social consensus under algorithmic quality control, it is not strictly equivalent to absolute factual truth and remains susceptible to coordinated rating manipulation. Moreover, trained annotators may share some of the same cultural or ideological assumptions as Community Notes contributors.

Third, our primary experiments use a gold-evidence setting to isolate adjudication from retrieval. We additionally conduct an end-to-end retrieval-enabled experiment, but the experiment is limited in scale. CoVer should therefore be viewed as complementary to retrieval systems.

Finally, some pairwise improvements on ContraNote Conflict benchmark do not reach significance. A power analysis suggests that approximately 4.9 times more sample are needed to detect the observed accuracy difference between CoVer and CONFACT at 80\% power, and approximately 2.4 times more samples for the comparison with ConflictRes. The current test set size limits the strength of our claims.

\section{Ethical Considerations}

The deployment of automated fact-checking systems involves potential ethical risks regarding information integrity. No automated system is infallible, and the risk of misclassification remains a primary concern. Incorrectly labeling a true claim as ``Refuted'' or a false claim as ``Supported'' can lead to the suppression of accurate information or the inadvertent spread of misinformation. Therefore, currently the CoVer framework should be treated as a decision-support tool for maintaining information integrity rather than an authority.

Regarding data privacy, ContraNote dataset is derived from the public \X Community Notes and acquired via the official \X API. We highlighted that reproduction or further use could be conducted with an \X API, so as to follow the official data usage terms. Furthermore, we emphasize that all research using such datasets must comply with the platform's terms of service, and respect the privacy and intent of the original content creators.


\bibliography{sample-base}

\appendix

\section{Generative AI Usage}

In accordance with generative AI usage policies, we disclose the use of Generative AI tools. We used generative AI as the base model for the experiment. Besides, we utilized Google's Gemini 3 Pro and ChatGPT (i.e., GPT-5.2) as writing assistants. Except for Figure \ref{fig:framework}, which was edited by Gemini Nano Banana, the Generative AI tool was used only for the purpose of improving the quality of writing. Its functions were limited to proofreading, language and clarity enhancement, conciseness, and word choice, and was not used to generate any core scientific content. The tool was applied to refine the final manuscript, after the content of each section was completed by the authors.

\section{Dataset Details}

\subsection{Comparisons of Datasets}\label{app:dataset}

Table~\ref{tab:dataset_positioning} contains the comparisons of datasets, where NEI denotes Not Enough Information. 

\begin{table*}[!htbp]
\centering
\small
\setlength{\tabcolsep}{3pt}
\resizebox{\textwidth}{!}{
\begin{tabular}{lcccccccccc}
\toprule
\textbf{Dataset}
& \textbf{Source domain}
& \textbf{Claim type}
& \textbf{Evidence type}
& \makecell{\textbf{Heterogeneous}\\\textbf{sources}}
& \makecell{\textbf{Explicit evidence}\\\textbf{conflict}}
& \makecell{\textbf{Subclaim}\\\textbf{decomposition}}
& \makecell{\textbf{Evidence-prioritization}\\\textbf{labels}}
& \makecell{\textbf{Social-media}\\\textbf{origin}}
& \makecell{\textbf{Claim-level}\\\textbf{labels}}
\\
\midrule

FEVER
& Wikipedia
& Factoid
& Wikipedia sentences
& No
& No
& No
& No
& No
& Supported / Refuted / NEI
\\

MultiFC
& Fact-checking websites
& Real-world claims
& Heterogeneous web documents
& Yes
& No
& No
& No
& Partly
& Claim-level veracity
\\

AVeriTeC
& Web and fact-checking sources
& Real-world claims
& Retrieved web evidence
& Yes
& No
& No
& No
& Partly
& Supported / Refuted
\\

ClaimDecomp
& Web fact-checking
& Complex claims
& Claim-linked evidence
& Yes
& No
& Yes
& No
& Partly
& Claim- and subclaim-level veracity
\\

WikiContradict
& Wikipedia
& Contradictory claims
& Wikipedia passages
& Limited
& Yes
& No
& No
& No
& Contradiction labels
\\

AmbiFC
& Real-world fact-checking
& Ambiguous claims
& Claim-linked evidence
& Yes
& Partly
& Partly
& No
& Partly
& Ambiguous / non-ambiguous
\\

ContraNote Conflict
& \X Community Notes
& Social-media claims
& Multiple user-generated notes
& Yes
& Yes
& Yes
& No
& Yes
& Supported / Refuted
\\

ContraNote Prioritization
& \X Community Notes
& Social-media claims
& Competing candidate notes
& Yes
& Yes
& Yes
& Yes
& Yes
& Supported-note / Refuted-note target
\\

\bottomrule
\end{tabular}
}
\caption{
Comparison of ContraNote with representative fact-checking datasets.
ContraNote Conflict provides claim-level verification labels, whereas ContraNote Prioritization additionally labels which competing evidence item should be prioritized.
}
\label{tab:dataset_positioning}
\end{table*}

\subsection{Dataset Analysis}\label{sec:dataset_analysis}

As shown in Figure~\ref{fig:comnote}, we provide an analysis of the \textbf{ContraNote} dataset. The primary reasons for flagging claims as misleading are missing important context (22,615 instances) and factual errors (20,918 instances). Other categories include unverified claims presented as facts (13,087 instances), outdated information (9,508 instances), satire (5,381 instances), manipulated media (5,124 instances), and other reasons (6,461 instances). Statistical analysis indicates that refuting notes are generally more detailed than supporting ones. Specifically, refuting notes average 1.77 per post and 289.08 characters in length, whereas supporting notes average 1.61 per post and 163.64 characters.

We grouped the cited evidence links in ContraNote into six source categories. Web and news pages constitute 49.1\% of the links, followed by social-media and video platforms (24.6\%), Wikipedia (8.6\%), government or other official sources (7.9\%), dedicated fact-checking sites (7.3\%), and other sources (2.5\%). The result shows that ContraNote is not dominated by short encyclopedic passages. A large share of its evidence comes from heterogeneous, socially situated sources for which authority and provenance need to be assessed.

Two researchers manually examined the semantic relation among a post, its strongest corrective note, and the competing note. The most frequent pattern is \textit{note-necessity disagreement}, accounting for 33\% of the manually coded cases. In these cases, the competing note does not necessarily establish that the post is factually true; instead, it argues that no Community Note is needed because the post is satire, opinion, or a platform-policy issue. The analysis also identifies \textit{evidential-authority disagreement}, in which notes rely on sources with different authority; \textit{scope or definition mismatch}, in which the same statement is evaluated under different temporal or definitional scopes; and \textit{entity/event attribution conflict}, in which evidence is attached to a different person, event, or provenance. For example, evidence may support a statement about a historical policy but fail to support the same statement when applied to current policy. These categories show that ContraNote contains contextual and aggregation-level disagreement in addition to direct factual negation. Because aggregate frequencies were not retained for remaining categories, we describe them qualitatively.

Furthermore, we analyzed the language distribution of posts with successfully retrieved text. Of these posts, 13,776 are in English, accounting for 62.98\%; 1,758 are in Spanish, accounting for 8.04\%; and 1,368 are in Portuguese, accounting for 6.25\%. Other posts are written in French, Japanese, German, Chinese, and other languages.

As compared in Table~\ref{tab:dataset_positioning}, ContraNote extends MultiFC~\cite{augenstein2019multifc}, AVeriTeC~\cite{schlichtkrull2023averitec}, and ClaimDecomp~\cite{chen2022generating}, which introduced real-world contrasting evidence into fact-checking. ContraNote specifically annotates naturally occurring evidence conflicts within the same post, and separates two forms of conflict (i.e., evidence- and aggregation-level). 

To assess the reliability of automatically derived labels, we randomly sampled 500 instances from ContraNote and recruited three trained annotators. They independently judged each post's veracity using the post content, the associated notes, and their cited evidence, achieving substantial inter-annotator agreement (Fleiss' $\kappa=0.82$). Majority vote human labels agreed with ContraNote labels on 94.2\% of instances. This validates labels' accuracy.

As Community Notes contributors may not represent all demographics, ContraNote may have population and ideological skew. We characterize its stance and topic distributions. Using PoliticalBiasBERT, we classify the notes into left, center, and right categories, with proportions of 18.6\%, 47.0\% and 34.4\% respectively. On the human-annotated subset, the topic distribution is politics/governance (26.0\%), health/ medicine (4.0\%), science/technology (20.6\%), media/entertainment/sports (29.8\%), economy/finance (5.8\%), public safety/crime (6.8\%), and other/ general (10.4\%). Topic annotations by three annotators achieved Fleiss' $\kappa=0.84$. These indicate that ContraNote reflects the consensus signal generated by Community Notes mechanism.

\section{Extended Evaluation}

\subsection{Class-wise Results for the Baseline Conditions}

We reported the class-wise results for the baseline conditions in Tables~\ref{tab:baseline_supported_prf} and~\ref{tab:baseline_refuted_prf}. Table~\ref{tab:baseline_supported_prf} showed the performance on the supported class, while Table~\ref{tab:baseline_refuted_prf} showed the performance on the refuted class.

\begin{table*}[t]
\centering
\resizebox{\textwidth}{!}{
\begin{tabular}{lccccccc}
\toprule
\textbf{Method} & \makecell[c]{\textbf{CONFACT}\\\textbf{HumC}} & \makecell[c]{\textbf{CONFACT}\\\textbf{ModC}} & \textbf{ConflictBank} & \textbf{ECON} & \textbf{FEVER} & \makecell[c]{\textbf{ContraNote}\\\textbf{Conflict}} & \makecell[c]{\textbf{ContraNote}\\\textbf{Prioritization}} \\
\midrule
FacTool & \res{38.5}{29.4}{33.3} & \res{57.6}{45.2}{50.7} & \res{31.9}{39.7}{35.4} & \res{71.1}{58.7}{64.3} & \res{90.7}{29.3}{44.3} & \res{34.7}{71.4}{46.7} & \res{56.6}{63.8}{60.0} \\
FactCheckGPT & \res{54.2}{38.2}{44.8} & \res{54.5}{42.9}{48.0} & \res{38.8}{44.8}{41.6} & \res{68.0}{55.4}{61.1} & \res{97.1}{74.4}{84.3} & \res{31.5}{48.6}{38.2} & \res{52.2}{64.5}{57.7} \\
FIRE & \res{30.0}{17.6}{22.2} & \res{48.6}{42.9}{45.6} & \res{27.1}{32.8}{29.7} & \res{62.2}{50.0}{55.4} & \res{93.3}{62.4}{74.8} & \res{34.4}{62.9}{44.4} & \res{64.6}{54.3}{59.0} \\
Confact & \res{34.8}{47.1}{40.0} & \res{53.6}{71.4}{61.2} & \res{37.0}{51.7}{43.2} & \res{68.9}{79.3}{73.7} & \res{88.5}{92.5}{90.4} & \res{51.1}{65.7}{57.5} & \res{59.4}{64.5}{61.9} \\
ConflictRes & \res{44.4}{35.3}{39.3} & \res{47.1}{57.1}{51.6} & \res{19.6}{15.5}{17.3} & \res{68.4}{58.7}{63.2} & \res{88.0}{71.4}{78.8} & \res{47.6}{57.1}{51.9} & \res{62.3}{45.7}{52.8} \\
\bottomrule
\end{tabular}
}
\caption{Detailed performance comparing different techniques, on the \textbf{Supported} class. Metrics are formatted as Precision $\mid$ \textcolor{blue}{Recall} $\mid$ \textcolor{purple}{F1-score}.}
\label{tab:baseline_supported_prf}
\end{table*}

\begin{table*}[t]
\centering
\resizebox{\textwidth}{!}{
\begin{tabular}{lccccccc}
\toprule
\textbf{Method} & \makecell[c]{\textbf{CONFACT}\\\textbf{HumC}} & \makecell[c]{\textbf{CONFACT}\\\textbf{ModC}} & \textbf{ConflictBank} & \textbf{ECON} & \textbf{FEVER} & \makecell[c]{\textbf{ContraNote}\\\textbf{Conflict}} & \makecell[c]{\textbf{ContraNote}\\\textbf{Prioritization}} \\
\midrule
FacTool & \res{86.2}{90.4}{88.2} & \res{86.2}{91.1}{88.6} & \res{72.4}{65.2}{68.7} & \res{69.4}{79.6}{74.1} & \res{40.1}{94.0}{56.3} & \res{92.0}{71.0}{80.1} & \res{63.4}{56.2}{59.6} \\
FactCheckGPT & \res{88.1}{93.4}{90.6} & \res{85.6}{90.5}{88.0} & \res{75.9}{71.1}{73.5} & \res{67.2}{77.8}{72.1} & \res{65.3}{95.5}{77.6} & \res{87.5}{77.3}{82.1} & \res{60.7}{48.1}{53.7} \\
FIRE & \res{84.4}{91.6}{87.9} & \res{85.2}{87.9}{86.5} & \res{69.8}{63.8}{66.7} & \res{63.5}{74.1}{68.4} & \res{55.0}{91.0}{68.5} & \res{90.3}{74.2}{81.5} & \res{64.5}{73.6}{68.7} \\
Confact & \res{88.2}{81.8}{84.9} & \res{91.5}{83.3}{87.2} & \res{76.5}{64.1}{69.7} & \res{79.8}{69.4}{74.3} & \res{83.6}{76.1}{79.7} & \res{92.2}{86.5}{89.2} & \res{66.3}{61.3}{63.7} \\
ConflictRes & \res{87.3}{91.0}{89.1} & \res{87.9}{82.9}{85.3} & \res{68.2}{73.9}{70.9} & \res{68.6}{76.9}{72.5} & \res{58.7}{80.6}{67.9} & \res{90.4}{86.5}{88.4} & \res{61.1}{75.5}{67.5} \\
\bottomrule
\end{tabular}
}
\caption{Detailed performance comparing different techniques, on the \textbf{Refuted} class. Metrics are formatted as Precision $\mid$ \textcolor{blue}{Recall} $\mid$ \textcolor{purple}{F1-score}.}
\label{tab:baseline_refuted_prf}
\end{table*}

\subsection{Additional Results on Baselines}\label{sec:additional}

We additionally evaluate ClaimDecomp, MADAM-RAG, and an AVeriTeC-style verification baseline in Table~\ref{tab:additional_baselines}. ClaimDecomp obtains mac. F1 scores of 67.28, 76.21, 54.29, and 65.58 on CONFACT-HumC, CONFACT-ModC, ContraNote Conflict, and ContraNote Prioritization, respectively. MADAM-RAG obtains 47.70 and 70.30 on the two ContraNote benchmarks, while the AVeriTeC-style baseline obtains 65.60 and 80.50. These results provide additional comparisons with methods designed for claim decomposition, retrieval-augmented verification, and conflict resolution.

On the additional datasets, CoVer obtains mac. F1 scores of 92.60 on AVeriTeC, 84.00 on WikiContradict, and 92.20 on AmbiFC. These experiments indicate that the framework is applicable beyond ContraNote, although the datasets differ in task formulation and evidence structure.

\begin{table*}[!htbp]
\centering
\small
\begin{tabular}{lcccc}
\toprule
\textbf{Method}
& \makecell{\textbf{CONFACT}\\\textbf{HumC}}
& \makecell{\textbf{CONFACT}\\\textbf{ModC}}
& \makecell{\textbf{ContraNote}\\\textbf{Conflict}}
& \makecell{\textbf{ContraNote}\\\textbf{Prioritization}}
\\
\midrule

MADAM-RAG
& 65.0 & 68.0 & 47.70 & 70.30
\\

AVeriTeC-style
& 65.8 & 77.4 & 65.60 & 80.50
\\

ClaimDecomp
& 67.28 & 76.21 & 54.29 & 65.58
\\

CoVer
& \textbf{77.10} & \textbf{83.10} & \textbf{68.00} & \textbf{88.50}
\\

\bottomrule
\end{tabular}
\caption{
Additional baseline comparisons using mac. F1 (\%). The AVeriTeC-style and MADAM-RAG results were reported for the ContraNote benchmarks, whereas ClaimDecomp was additionally evaluated on CONFACT. A dash denotes that the corresponding result was not reported.
}
\label{tab:additional_baselines}
\end{table*}

\subsection{Multi-Call Baseline}\label{sec:multicall}

Table~\ref{tab:shortcut_multicall} reports the shortcut and multi-call baselines. The metadata shortcut baseline achieves 81.5\% accuracy on ContraNote Conflict, but only 44.9 macro-F1 and 50.0 balanced accuracy. Its high accuracy is attributable to strong class imbalance rather than reliable verification. On ContraNote Prioritization, the same rule obtains 50.0\% accuracy, 33.3 macro-F1, and 50.0 balanced accuracy, which is close to chance. Thus, the benchmark cannot be adequately solved by the metadata rule alone.

For the matched-call control, we use fixed evidence for all four baselines within each task. We aggregate predictions by majority vote, breaking ties as Refuted, with results shown in Table~\ref{tab:shortcut_multicall}.

\begin{table}[!htbp]
\centering
\small
\begin{tabular}{lccc}
\toprule
\textbf{Method} & \textbf{Accuracy} & \textbf{Mac. F1} & \textbf{Bal. Acc.} \\
\midrule
\multicolumn{4}{l}{\textit{ContraNote Conflict}}\\
Helpful+misleading & 81.5 & 44.9 & 50.0 \\
CoVer & 86.0 & 68.0 & 64.5 \\
FactCheckGPT & 83.0 & 63.6 & 61.4 \\
CONFACT & 85.5 & 69.4 & 66.0 \\
FIRE & 85.5 & 67.4 & 63.9 \\
FacTool & 84.0 & 67.7 & 65.1 \\
\midrule
\multicolumn{4}{l}{\textit{ContraNote Prioritization}}\\
Helpful+misleading & 50.0 & 33.3 & 50.0 \\
CoVer & 88.5 & 88.5 & 89.2 \\
FactCheckGPT & 69.5 & 68.9 & 69.5 \\
CONFACT & 63.5 & 61.2 & 63.5 \\
FIRE & 61.0 & 58.4 & 61.0 \\
FacTool & 78.0 & 77.9 & 78.0 \\
\bottomrule
\end{tabular}
\caption{Shortcut and inference-budget controls on ContraNote. The shortcut rule predicts Refuted if and only if a candidate note is both helpful and misleading.}
\label{tab:shortcut_multicall}
\end{table}

\subsection{Three-way Verification}\label{app:three_way}

To examine whether CoVer can represent partial correctness, we construct a three-way setting with the labels Supported, Partially Supported, and Refuted. An instance is labeled Partially Supported when its subclaims contain both supported and unsupported or refuted components. Under this setting, CoVer obtains 80.6\% accuracy and 68.0 mac. F1, compared with 66.5 mac. F1 for the strongest baseline. This pilot experiment suggests that the framework can be extended beyond the binary formulation used by the primary ContraNote task.

\subsection{Factuality of Generated Rationales}\label{sec:factscore}

We evaluate the factuality of generated rationales using an adapted FActScore-style procedure. Atomic claims in the rationales are checked against the Community Notes and the associated evidence. We manually annotate 100 instances and use these annotations to validate the automatic procedure. GPT-5.5 agrees with the human annotations on 97\% of the cases. Under automatic annotation, CoVer achieves support rates of 85.2\% on ContraNote Conflict and 82.3\% on ContraNote Prioritization, compared with 76.9\% and 72.6\% for the strongest competing baselines, respectively.

\section{Ablation and Sensitivity Analysis}

\subsection{Class-wise Results for the Ablation Study}

We reported the class-wise results for the ablation study in Tables~\ref{tab:app_supported_prf} and~\ref{tab:app_refuted_prf}. Note that Table~\ref{tab:app_supported_prf} contains the detailed performance for the supported class, while Table~\ref{tab:app_refuted_prf} contains the detailed performance for the refuted class.

\begin{table*}[t]
\centering
\resizebox{\textwidth}{!}{
\begin{tabular}{lccccccc}
\toprule
\textbf{Setting} & \makecell[c]{\textbf{CONFACT}\\\textbf{HumC}} & \makecell[c]{\textbf{CONFACT}\\\textbf{ModC}} & \textbf{ConflictBank} & \textbf{ECON} & \textbf{FEVER} & \makecell[c]{\textbf{ContraNote}\\\textbf{Conflict}} & \makecell[c]{\textbf{ContraNote}\\\textbf{Prioritization}} \\
\midrule
Full & \res{72.0}{52.9}{61.0} & \res{77.8}{68.3}{72.7} & \res{56.8}{36.2}{44.2} & \res{73.3}{80.4}{76.7} & \res{98.4}{91.6}{94.9} & \res{73.3}{31.4}{44.0} & \res{80.3}{100.0}{89.1} \\
w/o Schema & \res{72.0}{52.9}{61.0} & \res{77.8}{68.3}{72.7} & \res{44.4}{27.6}{34.0} & \res{44.7}{16.3}{23.9} & \res{94.5}{78.0}{85.5} & \res{54.5}{51.4}{52.9} & \res{73.5}{79.8}{76.5} \\
w/o Consensus & \res{53.8}{61.8}{57.5} & \res{53.8}{68.3}{60.2} & \res{41.8}{65.5}{51.0} & \res{73.3}{80.4}{76.7} & \res{95.8}{86.5}{90.9} & \res{43.8}{20.0}{27.5} & \res{9.1}{2.1}{3.4} \\
w/o Support & \res{60.7}{50.0}{54.8} & \res{75.0}{65.9}{70.1} & \res{47.6}{34.5}{40.0} & \res{73.3}{80.4}{76.7} & \res{95.8}{86.5}{90.9} & \res{71.4}{28.6}{40.8} & \res{67.0}{64.9}{65.9} \\
w/o Schema + Consensus & \res{44.4}{47.1}{45.7} & \res{49.0}{61.0}{54.3} & \res{42.2}{65.5}{51.4} & \res{71.0}{82.6}{76.4} & \res{92.9}{88.7}{90.8} & \res{28.3}{74.3}{40.9} & \res{72.9}{45.7}{56.2} \\
w/o Schema + Support & \res{50.0}{26.5}{34.6} & \res{75.0}{51.2}{60.9} & \res{46.7}{36.2}{40.8} & \res{62.2}{30.4}{40.9} & \res{94.2}{85.0}{89.3} & \res{54.5}{51.4}{52.9} & \res{76.2}{81.9}{79.0} \\
w/o Consensus + Support & \res{53.8}{61.8}{57.5} & \res{53.8}{68.3}{60.2} & \res{41.8}{65.5}{51.0} & \res{73.3}{80.4}{76.7} & \res{95.8}{86.5}{90.9} & \res{53.3}{22.9}{32.0} & \res{4.0}{1.1}{1.7} \\
w/o All Three & \res{44.4}{47.1}{45.7} & \res{49.0}{61.0}{54.3} & \res{42.2}{65.5}{51.4} & \res{71.0}{82.6}{76.4} & \res{92.9}{88.7}{90.8} & \res{28.7}{77.1}{41.9} & \res{66.7}{40.4}{50.3} \\
\bottomrule
\end{tabular}
}
\caption{Detailed performance of the ablation study, on the \textbf{Supported} class. Metrics are formatted as Precision $\mid$ \textcolor{blue}{Recall} $\mid$ \textcolor{purple}{F1-score}.}
\label{tab:app_supported_prf}
\end{table*}

\begin{table*}[t]
\centering
\resizebox{\textwidth}{!}{
\begin{tabular}{lccccccc}
\toprule
\textbf{Setting} & \makecell[c]{\textbf{CONFACT}\\\textbf{HumC}} & \makecell[c]{\textbf{CONFACT}\\\textbf{ModC}} & \textbf{ConflictBank} & \textbf{ECON} & \textbf{FEVER} & \makecell[c]{\textbf{ContraNote}\\\textbf{Conflict}} & \makecell[c]{\textbf{ContraNote}\\\textbf{Prioritization}} \\
\midrule
Full & \res{90.8}{95.8}{93.2} & \res{92.0}{94.9}{93.5} & \res{77.3}{88.7}{82.6} & \res{81.8}{75.0}{78.3} & \res{85.5}{97.0}{90.9} & \res{87.0}{97.6}{92.0} & \res{100.0}{78.3}{87.8} \\
w/o Schema & \res{90.8}{95.8}{93.2} & \res{92.0}{94.9}{93.5} & \res{74.2}{85.8}{79.6} & \res{45.6}{77.7}{57.5} & \res{67.8}{91.0}{77.7} & \res{89.8}{90.9}{90.4} & \res{80.4}{74.3}{77.2} \\
w/o Consensus & \res{91.9}{89.1}{90.5} & \res{91.0}{84.5}{87.6} & \res{81.5}{62.4}{70.7} & \res{81.8}{75.0}{78.3} & \res{77.5}{92.5}{84.4} & \res{100.0}{79.4}{88.5} & \res{49.1}{78.3}{60.4} \\
w/o Support & \res{89.9}{93.3}{91.6} & \res{91.4}{94.3}{92.8} & \res{75.9}{84.5}{80.0} & \res{81.8}{75.0}{78.3} & \res{77.5}{92.5}{84.4} & \res{86.6}{97.6}{91.7} & \res{69.7}{71.7}{70.7} \\
w/o Schema + Consensus & \res{88.8}{87.7}{88.2} & \res{89.2}{83.5}{86.3} & \res{81.7}{63.1}{71.2} & \res{82.8}{71.3}{76.6} & \res{79.5}{86.6}{82.9} & \res{92.2}{57.0}{70.4} & \res{65.7}{61.3}{63.4} \\
w/o Schema + Support & \res{86.0}{94.4}{90.0} & \res{88.2}{95.5}{91.7} & \res{76.0}{83.0}{79.3} & \res{58.7}{84.3}{69.2} & \res{75.0}{89.6}{81.6} & \res{89.8}{90.9}{90.4} & \res{82.7}{77.1}{79.8} \\
w/o Consensus + Support & \res{91.9}{89.1}{90.5} & \res{91.0}{84.5}{87.6} & \res{81.5}{62.4}{70.7} & \res{81.8}{75.0}{78.3} & \res{77.5}{92.5}{84.4} & \res{98.6}{82.4}{89.8} & \res{47.9}{75.5}{58.6} \\
w/o All Three & \res{88.8}{87.7}{88.2} & \res{89.2}{83.5}{86.3} & \res{81.7}{63.1}{71.2} & \res{82.8}{71.3}{76.6} & \res{79.5}{86.6}{82.9} & \res{92.2}{57.6}{70.9} & \res{60.4}{58.1}{59.2} \\
\bottomrule
\end{tabular}
}
\caption{Detailed performance of the ablation study, on the \textbf{Refuted} class. Metrics are formatted as Precision $\mid$ \textcolor{blue}{Recall} $\mid$ \textcolor{purple}{F1-score}.}
\label{tab:app_refuted_prf}
\end{table*}

\subsection{Sensitivity to Quality Weights}\label{app:hyperparameters}

We evaluate the sensitivity of CoVer to the quality-weight vector $\boldsymbol{\lambda}=(\lambda_d,\lambda_a,\lambda_l,\lambda_u)$, corresponding to directness, attribute alignment, schema reliability, and informativeness, respectively. Each weight is varied over $\{0,0.05,0.10,\ldots,1.0\}$ subject to $\lambda_d+\lambda_a+\lambda_l+\lambda_u=1$. The uniform configuration $\boldsymbol{\lambda}_{\mathrm{uni}}=(0.25,0.25,0.25,0.25)$ is fixed before evaluation and is not tuned on any test set; we compare it with the best- and worst-performing configurations identified on the development split. All reported values are mac. F1 scores averaged over five runs. Across the evaluated datasets, mac. F1 varies within 1.3 percentage points, indicating that CoVer is not dependent on a narrowly tuned weighting configuration.

\begin{table*}[!htbp]
\centering
\small
\setlength{\tabcolsep}{3pt}
\begin{tabular}{lccccc}
\toprule
\textbf{Dataset}
& \makecell{\textbf{Uniform}\\$\boldsymbol{\lambda}$}
& \makecell{\textbf{Best}\\$\boldsymbol{\lambda}$}
& \makecell{\textbf{Worst}\\$\boldsymbol{\lambda}$}
& \makecell{\textbf{Best}\\\textbf{mac. F1}}
& \makecell{\textbf{Worst}\\\textbf{mac. F1}}
\\
\midrule

CONFACT-HumC
& 0.25,0.25,0.25,0.25
& 0.25,0.25,0.25,0.25
& 0.40,0.30,0.20,0.10
& 78.0
& 76.7
\\

CONFACT-ModC
& 0.25,0.25,0.25,0.25
& 0.25,0.25,0.25,0.25
& 0.10,0.20,0.30,0.40
& 84.0
& 82.7
\\

ConflictBank
& 0.25,0.25,0.25,0.25
& 0.25,0.25,0.25,0.25
& 0.40,0.30,0.20,0.10
& 64.3
& 63.0
\\

ECON
& 0.25,0.25,0.25,0.25
& 0.25,0.25,0.25,0.25
& 0.10,0.20,0.30,0.40
& 78.4
& 77.1
\\

FEVER
& 0.25,0.25,0.25,0.25
& 0.25,0.25,0.25,0.25
& 0.40,0.30,0.20,0.10
& 93.8
& 92.5
\\

ContraNote-Conflict
& 0.25,0.25,0.25,0.25
& 0.25,0.25,0.25,0.25
& 0.10,0.20,0.30,0.40
& 68.9
& 67.6
\\

ContraNote-Prioritization
& 0.25,0.25,0.25,0.25
& 0.25,0.25,0.25,0.25
& 0.40,0.30,0.20,0.10
& 89.4
& 88.1
\\

\bottomrule
\end{tabular}
\caption{
Sensitivity of CoVer to the quality-weight vector. Each weight vector is ordered as $(\lambda_d,\lambda_a,\lambda_l,\lambda_u)$. Mac. F1 values are reported in percentage points. The best and worst configurations are selected from the development split and evaluated without further tuning on the test set.
}
\label{tab:weight_sensitivity}
\end{table*}

\subsection{Tie-breaking Sensitivity}
\label{app:tiebreaking}

We compare the primary Refute-default rule in Eq.~\ref{eq:stance_tiebreak} with a Support-default variant. Changing the tie default decreases performance by 2.15 percentage points on ContraNote Prioritization but improves it by 4.19 percentage points on CONFACT-HumC. Thus, the effect of the tie-breaking rule is dataset-dependent rather than uniformly beneficial.

\subsection{Structured-Output Constraint Ablation}
\label{app:structured_constraints}

The factual-consensus call normally requires a complete structured record containing a stance from the valid label set and four quality scores bounded to $[0,1]$. We ablate these validity constraints while keeping the backbone model, evidence, and downstream decision rule unchanged; free-form outputs are parsed into the same intermediate fields whenever possible. Removing the structured-output constraints reduces accuracy by 1.57 percentage points relative to the constrained configuration. Thus, the constraints improve the stability of intermediate decisions, but the modest change indicates that CoVer's performance is not explained solely by output formatting.

\section{Efficiency and Error Analysis}

\subsection{Comparisons Controlling Inference Budget}\label{app:implementation}

Table~\ref{tab:inference_budget} showed the comparisons controlling inference budget.

\begin{table*}[!htbp]
\centering
\small
\setlength{\tabcolsep}{5pt}
\begin{tabular}{l l r r r r}
\toprule
\textbf{Method} &
\textbf{Configuration} &
\makecell{\textbf{Avg.}\\\textbf{LLM calls}} &
\makecell{\textbf{Avg. input}\\\textbf{tokens}} &
\makecell{\textbf{Avg. output}\\\textbf{tokens}} &
\makecell{\textbf{Estimated}\\\textbf{cost/request}} \\
\midrule
CoVer & Primary & 5.34 & 1274.9 & 271.2 & 0.0059 \\
FacTool & Single-call & 1.00 & 628.4 & 180.3 & 0.0034 \\
FacTool & Matched multi-call & 5.34 & 3261.5 & 852.2 & 0.0167 \\
FactCheckGPT & Single-call & 1.00 & 635.4 & 357.1 & 0.0052 \\
FactCheckGPT & Matched multi-call & 5.34 & 3298.2 & 1461.2 & 0.0229 \\
FIRE & Single-call & 1.00 & 633.4 & 275.1 & 0.0043 \\
FIRE & Matched multi-call & 5.34 & 3288.0 & 1530.4 & 0.0235 \\
CONFACT & Single-call & 1.00 & 632.4 & 196.0 & 0.0035 \\
CONFACT & Matched multi-call & 5.34 & 3282.7 & 938.0 & 0.0176 \\
ConflictRes & Single-call & 1.00 & 632.4 & 188.5 & 0.0035 \\
MADAM-RAG & Single-call & 1.00 & 571.4 & 173.8 & 0.0032 \\
AVeriTeC-style & Single-call & 1.00 & 572.4 & 155.1 & 0.0030 \\
ClaimDecomp & Single-call & 1.00 & 635.4 & 341.6 & 0.0050 \\
\bottomrule
\end{tabular}
\caption{
Inference budget and computational cost under the original and matched-call configurations. Reported values are averages per instance over all evaluation examples. Input and output tokens include all LLM requests made by a method. The estimated cost is calculated using the API prices corresponding to the reported backbone model and evaluation date.
}
\label{tab:inference_budget}
\end{table*}

\subsection{Error Analysis}\label{app:error_analysis}

We qualitatively inspect CoVer's incorrect predictions to identify recurring failure modes. The analysis reveals three principal categories.

\textbf{Temporal ambiguity.}
Some claims and notes refer to different stages of an evolving event or omit the relevant time frame. In such cases, evidence that was correct at one point can conflict with a later update, and the model may select an outdated interpretation or fail to restrict the verdict to the claim's intended period.

\textbf{Evidence requiring domain expertise.}
Some cases depend on specialized legal, medical, scientific, or policy knowledge that is not stated explicitly in the supplied evidence. CoVer can identify the competing stances but may assign excessive weight to a fluent explanation when resolving the technical distinction requires expert interpretation.

\textbf{True event with an unsupported implication.}
A claim may mention a real event and then attach an unsupported causal, intentional, or generalized implication. The model sometimes treats evidence for the underlying event as support for the entire claim, even when the implication is not entailed. This failure mode motivates the conservative support-verification stage, but difficult cases remain when the factual and implied components are tightly coupled.

These errors suggest three corresponding directions for improvement: explicit temporal normalization, routing of specialized cases to domain-aware evidence or experts, and finer-grained decomposition of event facts from causal or intentional implications. This analysis is qualitative; we do not assign category percentages because the available coding record does not contain a frequency table.

\end{document}